\documentclass[preprint,12pt]{elsarticle}

\usepackage{amssymb}
\usepackage{amsmath}
\usepackage{algorithm}
\usepackage{algpseudocode}

\journal{Nuclear Physics B}

\begin{document}

\begin{frontmatter}



\title{Title: A Two-Stage Detection–Tracking Framework for Stable Apple Quality Inspection in Dense Conveyor-Belt Environments
}


\author[label1]{Keonvin Park\fnref{fn1}}
\ead{kbpark16@snu.ac.kr}

\author[label2]{Aditya Pal\fnref{fn1}}
\ead{aditya@dgu.ac.kr}

\author[label3]{Jin Hong Mok\corref{cor1}}
\ead{jhmok1024@dgu.edu}

\fntext[fn1]{These authors contributed equally to this work.}

\cortext[cor1]{Corresponding author. Email: jhmok1024@dgu.edu}

\affiliation[label1]{organization={Interdisciplinary Program in Artificial Intelligence, Seoul National University},
            addressline={1 Gwanak-ro, Gwanak-gu},
            city={Seoul},
            postcode={08826},
            country={Republic of Korea}}

\affiliation[label2]{organization={Department of Biological Environmental Science, College of Life Science and Biotechnology, Dongguk University},
            city={Seoul},
            postcode={04620},
            country={Republic of Korea}}

\affiliation[label3]{organization={ Department of Food Science and Biotechnology, Dongguk University},
            city={Goyang-si, Gyeonggi-do},
            postcode={10326},
            country={Republic of Korea}}

\begin{abstract}
Industrial fruit inspection systems must operate reliably under dense multi-object interactions and continuous motion, yet most existing works evaluate detection or classification at the image level without ensuring temporal stability in video streams. We present a two-stage detection–tracking framework for stable multi-apple quality inspection in conveyor-belt environments. An orchard-trained YOLOv8 model performs apple localization, followed by ByteTrack multi-object tracking to maintain persistent identities. A ResNet18 defect classifier, fine-tuned on a healthy–defective fruit dataset, is applied to cropped apple regions. Track-level aggregation is introduced to enforce temporal consistency and reduce prediction oscillation across frames. We define video-level industrial metrics such as track-level defect ratio and temporal consistency to evaluate system robustness under realistic processing conditions. Results demonstrate improved stability compared to frame-wise inference, suggesting that integrating tracking is essential for practical automated fruit grading systems.
\end{abstract}

\begin{graphicalabstract}
\includegraphics[width=\textwidth]{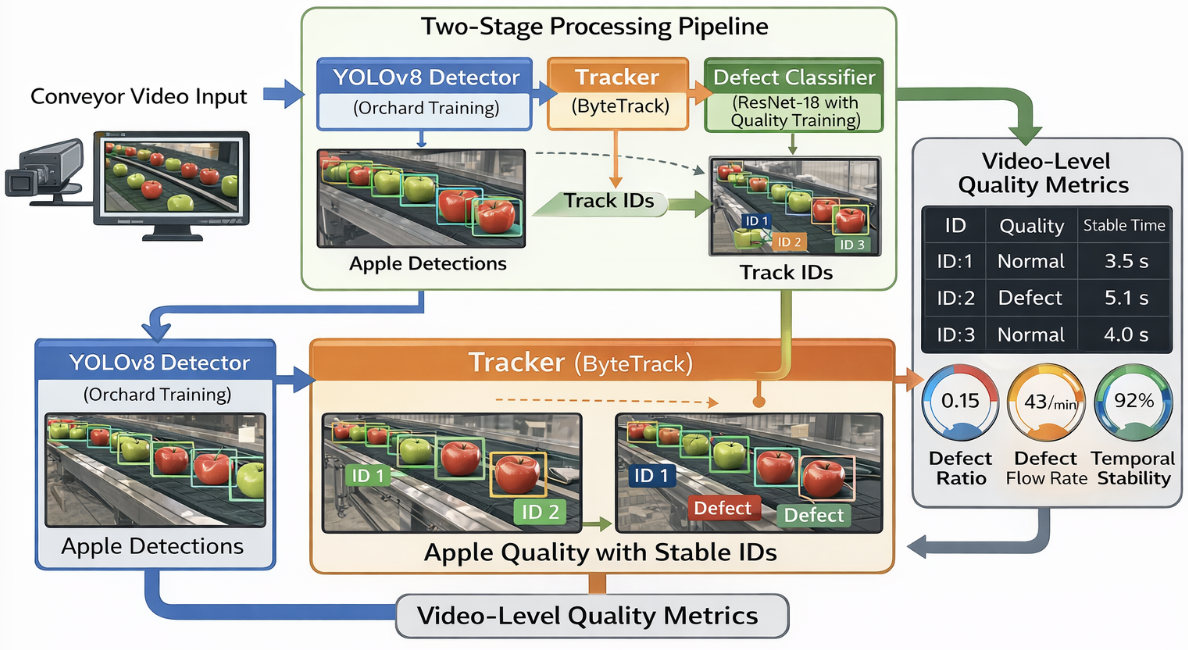}
\end{graphicalabstract}

\begin{highlights}
\item Proposes a preliminary two-stage detection--tracking framework for automated apple quality inspection in dense conveyor-belt environments.
\item Integrates an orchard-trained YOLOv8 detector with ByteTrack to ensure persistent object identities across video frames.
\item Employs a ResNet18-based defect classifier, fine-tuned on a healthy--defective fruit dataset, for per-apple quality prediction.
\item Introduces track-level aggregation to stabilize defect decisions and mitigate frame-wise prediction inconsistency.
\item Defines video-level industrial quality metrics, including track-level defect ratio and temporal stability, for conveyor-based evaluation.
\end{highlights}


\end{frontmatter}



\section{Introduction}
\label{sec1}

Automated fruit quality inspection plays a critical role in modern agricultural processing and grading systems. In industrial conveyor-belt environments, apples are densely arranged and continuously transported, requiring real-time, stable, and reliable quality assessment. Traditional manual inspection is labor-intensive and subjective, motivating the development of computer vision-based grading systems.

Recent advances in deep learning have significantly improved object detection performance, with models such as YOLO achieving high accuracy and real-time speed in diverse scenarios \cite{Redmon2016, YOLOv8}. In agricultural contexts, fruit detection has been extensively studied in orchard environments, where models are trained to localize fruits under natural lighting and background conditions \cite{Sa2016, Chen2021}. However, most existing works focus primarily on image-level detection or classification, without explicitly addressing object-level temporal consistency in video streams.

In industrial settings, quality inspection requires not only detecting apples but also ensuring stable per-object decisions across consecutive frames. Frame-wise classification can produce fluctuating predictions due to motion blur, occlusion, and illumination variation, leading to unreliable grading outcomes. Multi-object tracking algorithms such as ByteTrack \cite{ByteTrack2022} enable persistent identity assignment, offering a mechanism to enforce object-level temporal reasoning.

In parallel, defect classification for fruit grading has been widely explored using convolutional neural networks \cite{Mohanty2016, FruitDefectSurvey}. Nevertheless, most approaches evaluate classification performance on static image datasets, without integration into a video-based multi-object pipeline. The gap between image-level benchmarks and video-level industrial deployment remains underexplored.

To address these limitations, we propose a two-stage detection--tracking framework for stable apple quality inspection in dense conveyor-belt environments. First, an orchard-trained YOLOv8 detector localizes apples in industrial conveyor videos. Second, ByteTrack is applied to assign persistent identities across frames. For each tracked apple, a ResNet18-based defect classifier, initialized with ImageNet weights and fine-tuned on a healthy--defective fruit dataset, predicts quality categories. Track-level aggregation is employed to stabilize defect decisions and mitigate frame-wise prediction oscillation.

Furthermore, we introduce video-level industrial quality metrics, including track-level defect ratio and temporal consistency, to evaluate system performance beyond conventional image-based accuracy. Our preliminary study demonstrates that incorporating tracking significantly improves decision stability compared to frame-wise classification, highlighting the importance of temporal modeling for practical industrial fruit grading systems.

\section{Related Work}
\label{sec:related}

\subsection{Fruit Detection in Agricultural Environments}

Deep learning-based fruit detection has been extensively studied in orchard settings. Early approaches employed region-based convolutional neural networks for fruit localization \cite{Sa2016}. With the emergence of one-stage detectors such as YOLO \cite{Redmon2016}, real-time fruit detection has become feasible in field robotics and harvesting applications. Recent works leverage improved YOLO variants and attention mechanisms for enhanced robustness \cite{Koirala2019}. However, these approaches predominantly operate on static images and do not address object-level temporal consistency in industrial conveyor environments.

\subsection{Multi-Object Tracking}

Multi-object tracking (MOT) aims to associate detections across frames to maintain consistent identities. Classical evaluation metrics such as CLEAR MOT \cite{MOTMetrics} quantify tracking performance. Recent tracking-by-detection paradigms have achieved strong results by associating detection boxes across frames. ByteTrack \cite{ByteTrack2022} improves tracking robustness by associating both high- and low-confidence detections, making it suitable for dense and dynamic scenes. While MOT has been widely adopted in surveillance and autonomous driving, its integration into industrial fruit grading remains limited.

\subsection{Fruit Defect and Quality Classification}

Defect detection and quality assessment of fruits have been explored using convolutional neural networks and transfer learning strategies. Mohanty et al.~\cite{Mohanty2016} demonstrated the effectiveness of deep CNNs for plant disease classification. Surveys in agricultural vision highlight defect classification as a key task in post-harvest management \cite{FruitDefectSurvey}. Nevertheless, most prior works evaluate classification models on curated static datasets, without considering dense multi-object video scenarios or track-level temporal stabilization.

\subsection{Domain Shift and Industrial Deployment Gap}

Models trained in orchard environments often experience performance degradation when deployed in processing facilities due to domain shift \cite{Pan2010, DGSurvey2022}. Differences in lighting, background, object arrangement, and motion dynamics introduce distribution shifts that challenge generalization. Despite extensive research on domain adaptation and generalization, few studies examine how image-level agricultural models behave when deployed in industrial conveyor-belt videos.

Our work bridges these gaps by integrating detection, tracking, and defect classification into a unified video-based quality inspection framework, explicitly addressing temporal stability and industrial evaluation metrics.

\section{Methods}

We propose a two-stage detection--tracking framework for stable apple quality inspection in conveyor-belt videos. The system consists of (1) apple detection using YOLOv8, (2) multi-object tracking via ByteTrack, and (3) defect classification using a ResNet18 model with track-level aggregation for temporal stabilization.

\subsection{Apple Detection}

Given an input video consisting of frames $\{F_t\}_{t=1}^{T}$, we first detect apples in each frame using a pretrained YOLOv8 detector \cite{YOLOv8}. The detector produces a set of bounding boxes:

\begin{equation}
\mathcal{D}_t = \{b_t^1, b_t^2, \dots, b_t^{N_t}\}
\end{equation}

where $N_t$ denotes the number of detected apples at frame $t$.

\subsection{Multi-Object Tracking}

To maintain persistent object identities across frames, we apply ByteTrack \cite{ByteTrack2022}. The tracker associates detections across consecutive frames and assigns a unique track ID $i$ to each apple instance. This produces track trajectories:

\begin{equation}
\mathcal{T}_i = \{b_{t_1}^i, b_{t_2}^i, \dots, b_{t_k}^i\}
\end{equation}

where $k$ denotes the track length.

Tracking enables object-level reasoning and prevents frame-wise identity switching in dense conveyor environments.

\subsection{Defect Classification}

For each detected bounding box, we crop the corresponding apple region and classify its quality using a ResNet18 model \cite{He2016}, initialized with ImageNet weights and fine-tuned on the Healthy-Defective dataset \cite{HealthyDefectiveDataset}.

Given a cropped region $x_t^i$, the classifier outputs a predicted defect class:

\begin{equation}
y_t^i = f_{\text{ResNet18}}(x_t^i)
\end{equation}

where $y_t^i \in \{0, 1, \dots, C-1\}$ represents the defect category.

\subsection{Track-Level Aggregation}

Frame-wise predictions may fluctuate due to motion blur, occlusion, or illumination changes. To enforce temporal consistency, we aggregate predictions across each track.

For track $i$, we collect all predictions:

\begin{equation}
\mathcal{Y}_i = \{y_{t_1}^i, y_{t_2}^i, \dots, y_{t_k}^i\}
\end{equation}

We apply majority voting to determine the final track-level label:

\begin{equation}
\hat{y}_i = \arg\max_{c} \sum_{t} \mathbb{I}(y_t^i = c)
\end{equation}

where $\mathbb{I}(\cdot)$ is the indicator function.

This aggregation stabilizes defect decisions and reduces frame-level oscillation.

\subsection{Overall Pipeline}

The full procedure is summarized in Algorithm~\ref{alg:twostage}.

\begin{algorithm}[H]
\caption{Two-Stage Detection--Tracking--Defect Pipeline}
\label{alg:twostage}
\begin{algorithmic}[1]
\Require Video frames $\{F_t\}_{t=1}^{T}$
\Ensure Track-level quality labels $\{\hat{y}_i\}$

\For{each frame $F_t$}
    \State Detect apples using YOLOv8
    \State Associate detections using ByteTrack to obtain track IDs
    \For{each detected apple with track ID $i$}
        \State Crop apple region
        \State Predict defect class using ResNet18
        \State Store prediction in memory buffer for track $i$
    \EndFor
\EndFor

\For{each track $i$}
    \State Compute majority voting over stored predictions
    \State Assign final quality label $\hat{y}_i$
\EndFor
\end{algorithmic}
\end{algorithm}

\section{Data}

Our framework is trained using two image-level datasets and evaluated on industrial conveyor-belt video data.

\subsection{Orchard-Based Apple Detection Dataset}

For apple localization, we adopt an orchard-trained YOLOv8 detector \cite{YOLOv8}. The detector is pretrained on an apple detection dataset collected in orchard environments, where apples are annotated with bounding boxes under natural lighting conditions. Similar orchard-based fruit detection datasets have been widely used in agricultural robotics research \cite{Sa2016}. The trained detector is directly deployed in industrial conveyor scenes without additional fine-tuning, allowing us to evaluate cross-domain robustness.

\subsection{Healthy--Defective Apple Dataset}

For defect classification, we use the Healthy-Defective Fruits dataset \cite{HealthyDefectiveDataset}, which provides real apple images categorized into \textit{fresh}, \textit{bruise\_defect}, \textit{rot\_defect}, and \textit{scab\_defect}. The dataset contains real-world fruit images acquired under controlled conditions and is suitable for supervised defect learning.

We construct train/validation/test splits using a fixed random seed to ensure reproducibility. A ResNet18 model \cite{He2016}, initialized with ImageNet weights, is fine-tuned on this dataset for multi-class defect classification. During inference, defect classes are aggregated into binary labels (normal vs. defect) for industrial evaluation.

\subsection{Industrial Conveyor Video Data}

To evaluate the proposed framework under realistic processing conditions, we use publicly available conveyor-belt apple videos obtained from YouTube. These videos depict densely arranged apples transported on industrial conveyor systems, exhibiting motion blur, partial occlusion, and illumination variations.

Because no frame-level defect annotations are available, the videos are used exclusively for inference and video-level evaluation. This setup reflects a preliminary feasibility study designed to analyze temporal stability and object-level decision consistency under domain shift.

\section{Experiments}

\subsection{Experimental Setup}

The proposed system consists of three components: (1) YOLOv8-based apple detection, (2) ByteTrack multi-object tracking \cite{ByteTrack2022}, and (3) ResNet18-based defect classification.

The YOLOv8 detector is pretrained on orchard data and directly applied to conveyor videos. ByteTrack associates detection boxes across frames to maintain persistent track identities. For each tracked apple, cropped image regions are passed to a ResNet18 classifier fine-tuned on the Healthy-Defective dataset.

\subsection{Evaluation Metrics}

\paragraph{Detection Metrics.}
Detection performance is evaluated using mean Average Precision (mAP) and precision–recall statistics.

\paragraph{Classification Metrics.}
Defect classification performance on the held-out test set is evaluated using accuracy, precision, recall, and F1-score.

\paragraph{Video-Level Industrial Metrics.}
Since frame-level annotations are unavailable for conveyor videos, we define track-level quality metrics:

\begin{equation}
\text{Defect Ratio} = \frac{N_{\text{defect tracks}}}{N_{\text{total tracks}}}
\end{equation}

\begin{equation}
\text{Temporal Stability} = 1 - \frac{\text{Number of label changes per track}}{\text{Track length}}
\end{equation}

Track-level majority voting is applied to reduce frame-wise prediction fluctuations.

\subsection{Implementation Details}

The defect classifier is trained using Adam optimizer with learning rate $1e^{-4}$ and batch size 32 for 10 epochs. All experiments are conducted using PyTorch. Inference is performed on Apple MPS backend.

\section{Data}

Our framework is trained using two image-level datasets and evaluated on industrial conveyor-belt video data.

\subsection{Orchard-Based Apple Detection Dataset}

For apple localization, we adopt an orchard-trained YOLOv8 detector \cite{YOLOv8}. The detector is pretrained on an apple detection dataset collected in orchard environments, where apples are annotated with bounding boxes under natural lighting conditions. Similar orchard-based fruit detection datasets have been widely used in agricultural robotics research \cite{Sa2016}. The trained detector is directly deployed in industrial conveyor scenes without additional fine-tuning, allowing us to evaluate cross-domain robustness.

\subsection{Healthy--Defective Apple Dataset}

For defect classification, we use the Healthy-Defective Fruits dataset \cite{HealthyDefectiveDataset}, which provides real apple images categorized into \textit{fresh}, \textit{bruise\_defect}, \textit{rot\_defect}, and \textit{scab\_defect}. The dataset contains real-world fruit images acquired under controlled conditions and is suitable for supervised defect learning.

We construct train/validation/test splits using a fixed random seed to ensure reproducibility. A ResNet18 model \cite{He2016}, initialized with ImageNet weights, is fine-tuned on this dataset for multi-class defect classification. During inference, defect classes are aggregated into binary labels (normal vs. defect) for industrial evaluation.

\subsection{Industrial Conveyor Video Data}

To evaluate the proposed framework under realistic processing conditions, we use publicly available conveyor-belt apple videos obtained from YouTube. These videos depict densely arranged apples transported on industrial conveyor systems, exhibiting motion blur, partial occlusion, and illumination variations.

Because no frame-level defect annotations are available, the videos are used exclusively for inference and video-level evaluation. This setup reflects a preliminary feasibility study designed to analyze temporal stability and object-level decision consistency under domain shift.

\section{Expected Results}

As a preliminary feasibility study, we expect the proposed two-stage detection--tracking framework to demonstrate improved temporal stability and object-level consistency compared to frame-wise quality classification.

\subsection{Detection Performance}

The orchard-trained YOLOv8 detector is expected to achieve high detection precision in conveyor scenes, although minor performance degradation may occur due to domain shift between orchard and industrial environments.

\subsection{Defect Classification Performance}

The ResNet18-based defect classifier, fine-tuned on the Healthy-Defective dataset, is expected to achieve strong image-level classification performance on the held-out test set. However, direct frame-wise deployment in conveyor videos may produce prediction fluctuations caused by motion blur, occlusion, and lighting variation.

\subsection{Effect of Tracking and Aggregation}

By incorporating ByteTrack and track-level majority voting, we expect:

\begin{itemize}
    \item Reduced prediction oscillation across frames.
    \item Improved track-level defect decision stability.
    \item More reliable defect ratio estimation at the video level.
\end{itemize}

We hypothesize that track-level aggregation will significantly improve temporal stability compared to frame-wise classification without tracking.

\section{Conclusion}
\label{sec:conclusion}

We introduced a detection--tracking--classification framework for stable multi-apple quality inspection in conveyor-belt videos. By combining YOLOv8 detection, ByteTrack identity association, and ResNet18-based defect classification with track-level aggregation, the proposed system addresses temporal instability commonly observed in frame-wise industrial inspection.

Our preliminary study demonstrates that enforcing object-level temporal consistency significantly improves the reliability of defect estimation in dense and dynamic scenes. The proposed video-level evaluation metrics further bridge the gap between image-based research benchmarks and real-world industrial deployment.

These findings indicate that tracking-aware quality inspection is essential for practical fruit grading systems. Future work will investigate domain generalization techniques, multi-camera integration, and large-scale industrial validation.


\end{document}